\documentclass{article}
\usepackage{amsmath}
\usepackage{graphicx}
\usepackage{float}
\usepackage{hyperref}
\title{Hybrid Training Approaches for LLMs: Leveraging Real and Synthetic Data to Enhance Model Performance in Domain-Specific Applications}
\author{Alexey Zhezherau M.Mgmt,  Alexei Yanockin MSc}
\date{Oct 7, 2024}

\begin{document}

\maketitle

\begin{abstract}
This research explores a hybrid approach to fine-tuning large language models (LLMs) by integrating real-world and synthetic data, with the goal of enhancing model performance, particularly in generating accurate and contextually relevant responses. Leveraging a dataset that combines transcribed real interactions with high-quality synthetic sessions, we aimed to overcome the limitations of scarce, noisy, and domain-specific real data by using synthetic personas and scenarios to enrich training diversity. The study evaluated three models: a base foundational model, a model fine-tuned with real data, and a hybrid fine-tuned model. Experimental results demonstrated that the hybrid model consistently outperformed the others in specific vertical applications, achieving the highest scores across all metrics. Further testing confirmed the hybrid model's superior adaptability and contextual understanding across diverse scenarios. The findings indicate that combining real and synthetic data offers significant advantages in improving the robustness and contextual sensitivity of LLMs, especially in domain-specific and vertical use cases.


\end{abstract}

\section{Introduction}
The introduction of the Transformer architecture, as presented in the foundational paper \textit{Attention Is All You Need} \cite{vaswani2017attention}, fundamentally reshaped the landscape of natural language processing (NLP) and generative AI by shifting away from traditional recurrent and convolutional architectures. Transformers introduced the concept of self-attention mechanisms, which allowed models to weigh the relevance of different words in a sequence simultaneously, leading to better contextual understanding and parallelizability in training. This innovation provided the foundation for subsequent breakthroughs in large language models (LLMs), such as OpenAI's GPT, Google's BERT, and open-source players, which marked a new era in language understanding and generation.

Since the introduction of Transformers, significant advancements have continued to shape the field. Generative Pre-trained Transformers (GPT) models, especially GPT-3, demonstrated that scaling model size and data volume could significantly enhance language capabilities \cite{xu2023peft}. This scaling law, alongside innovations such as sparse attention mechanisms and parameter-efficient fine-tuning, allowed models like LLaMA and ChatGPT to further push the boundaries of what is achievable in NLP.

AI's progress is driven by three pillars: compute, foundational model advancements, and data \cite{nopriors2023podcast}. Compute power enables AI to process vast amounts of information, while cutting-edge algorithms provide new learning techniques and methods to optimize performance. However, the competitive edge for companies increasingly lies in data ownership and quality. Data is the key resource that fuels AI's learning and adaptability across various scenarios. Leading companies have already exhausted publicly available web data, and this approach is reaching its limits. The next step involves accessing unique, high-quality datasets and finding innovative ways to improve training data through hybrid methods.

This research aims to address the following question: Can we accelerate model fine-tuning and improve large language models ability to generate accurate, contextually appropriate responses by fine-tuning it with both real-world and synthetic data? The hypothesis is that this hybrid approach will outperform models trained solely on real-world data or those relying on sophisticated prompt engineering with a base model. By integrating synthetic data, the aim is to make the model more versatile and capable of generalizing across diverse contexts.

High-quality, diverse real-world data for specific use cases is often scarce, expensive to collect, and may not cover all scenarios required for domain-specific applications \cite{nikolenko2019synthetic}. By supplementing real data with synthetic data, we may be able to fine-tune AI models more effectively and efficiently, ensuring that the resulting models can handle both common and edge-case scenarios.

In this paper, we describe the development of a unique training dataset and the instruction-based fine-tuning of a large language model. We use the CBT Therapy Counselor project as an example, but the principles apply to numerous fields. By demonstrating how synthetic data can fill gaps left by real data, we hope to show the broader utility of hybrid training strategies in advancing large language models.

\section{Why the Future of LLM Training might be Combining Real and Synthetic Data}
Real-world data is invaluable but has limitations: it is often messy, incomplete, and narrow in scope. Synthetic data can address some of these challenges \cite{guo2024generative}:
\begin{itemize}
    \item \textbf{Scalability:} Generate large volumes of diverse data rapidly. Synthetic data generation allows for quick creation of relevant training examples, enabling a more efficient training process \cite{chung2022scaling} \cite{chan2024balancing}.
    \item \textbf{Customization:} Create scenarios tailored to specific use cases. Synthetic data can be designed to replicate rare or important scenarios that may not occur frequently enough in real datasets.
    \item \textbf{Edge Cases:} Produce examples of rare but critical situations. Addressing these edge cases is crucial for models that need to be robust and capable of handling unexpected or uncommon inputs.
\end{itemize}

This approach is particularly valuable in domain-specific applications where exposing the model to a wide range of specific scenarios and interactions is crucial \cite{lhoest2021datasets}. Recent work by \cite{qin2023read} has emphasized the importance of explainable LLMs, particularly in sensitive contexts such as mental health, which benefits from the diverse exposure that hybrid training can provide. Similarly, \cite{chen2023chatbots} have illustrated how LLM-powered chatbots for psychiatric simulations can be improved through exposure to both real and synthetic conversational data. Such hybrid approaches enrich the model's ability to understand nuances that purely real data may not provide due to inherent limitations \cite{chung2022scaling}.

Our experiments indicate that this hybrid approach consistently outperformed models trained solely on real-world data and those that relied solely on sophisticated prompt engineering with a base model. The use of synthetic data as a complementary training tool—especially in sensitive domains such as mental health, as discussed in \cite{jin2023psyeval} — allowed us to expand the model's capabilities, making it more flexible and better suited to handling complex, varied inputs. Moreover, \cite{hua2024llms} have highlighted the increasing relevance of synthetic data in mental health care, suggesting that synthetic sessions can serve as effective supplements when real-world examples are either scarce or ethically challenging to obtain.

\section{Data Strategy: Building a 500-Session Training Dataset}

\subsection{Real Data: 300+ Scraped and Transcribed Counseling Sessions}
To develop the AI Therapy Counselor, over 300 actual counseling sessions were collected, which provided a substantial dataset for model training and evaluation. The sessions encompassed a variety of counseling techniques, topics, and different approaches to client interaction, allowing the AI to be exposed to a diverse range of real-world counseling examples.

\subsubsection{Data Source}
CBT therapy session role-play transcriptions were scraped from a well-known video hosting platform, providing diverse, publicly available therapy simulations. These sessions covered a broad range of topics, including stress management, anxiety, relationship issues, and personal growth, ensuring that the data had enough variety to serve as a strong foundation for model training. This variety was crucial in ensuring the model's ability to handle a wide spectrum of conversational themes and challenges.

\subsubsection{Data Processing}
The transcriptions underwent rigorous preprocessing inspired by data preparation techniques described in foundational researches \cite{devlin2019bert} and \cite{liu2019roberta}. This involved segmentation, and filtering steps to ensure consistency and dataset quality. Identifying information, noise, and redundant elements were meticulously removed, and dialogues were segmented into question-answer pairs, much like the comprehensive data preprocessing strategies employed by RoBERTa to enhance model robustness.

Each dialogue was further enriched through the analysis of key conversational cues, therapist interventions, and emotional indicators, leveraging insights from \cite{raffel2019t5}. These annotations facilitated more nuanced learning during the training phase, enabling the model to understand not only the content but also the contextual and emotional dynamics of conversations, ultimately improving downstream performance. Techniques for deduplication and quality filtering from \cite{brown2020language} GPT-3 work (2020) were also integrated to eliminate redundancy and ensure data reliability, providing a high-quality foundation for fine-tuning.

To mitigate biases and reduce noise, extensive preprocessing and filtering were applied, as outlined in the EleutherAI research \cite{biderman2020eleutherai}. This ensured that the dataset was of high quality, anchoring the AI model in realistic scenarios that set the groundwork for subsequent synthetic data augmentation. By grounding the model in meticulously cleaned and authentic real-world data, we provided an initial level of context-specific knowledge and authenticity, which was further enhanced through synthetic sessions. This foundational approach allowed the model to build a broader understanding through synthetic augmentation, leading to an AI system that is both versatile and deeply informed by real-world context.

\subsection{Synthetic Persona Generation}
To address the limitations of real-world data, a set of synthetic personas and counseling themes was created. These imaginary clients were diverse in backgrounds, personalities, and issues, ensuring the AI learned to interact with a wide variety of people. The diversity of personas allowed the model to generalize better across different types of users and situations, making it more adaptable and flexible.

The flow chart below gives an overview of how synthetic therapy sessions were created, showing the steps involved in generating and refining the data.
    
 \begin{figure}[H]
    \centering    
 	\includegraphics [width=\linewidth] {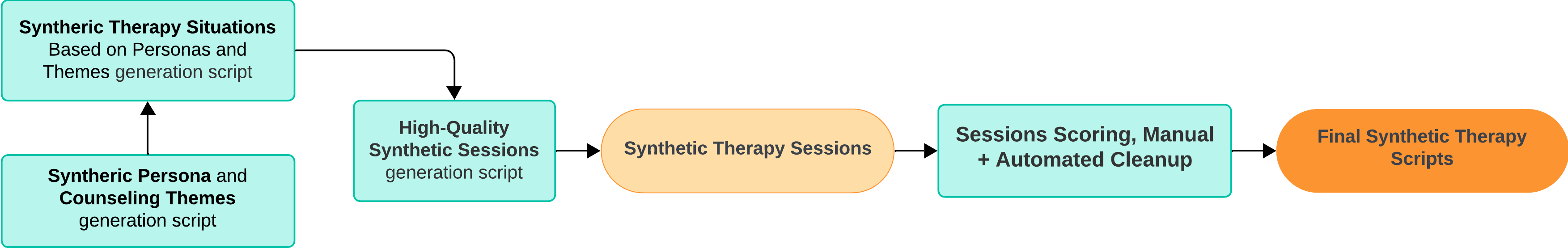}
 	
    \caption{Flowchart showing the steps involved in generating and refining synthetic therapy sessions.}
    \label{fig:generating_and_refining}
\end{figure}

Drawing from research insights in ZeroGen \cite{ye2022zerogen}, the quality of synthetic personas and overall synthetic data generation was also evaluated based on metrics such as diversity, correctness, and naturalness. Diversity was a critical consideration, ensuring that the generated personas differed significantly in terms of demographic and psychological characteristics. This diversity allowed the model to encounter a wide spectrum of scenarios, similar to those observed in real-world counseling. Correctness was assessed to ensure that the synthetic personas and their corresponding scenarios accurately reflected the intended themes and therapeutic contexts. Naturalness was measured to confirm that the interactions and characteristics of the personas were plausible and resembled real human behavior, thereby enhancing the realism of the training data.

Each synthetic persona was designed with specific traits, including age, occupation, cultural background, and unique personal challenges. These elements were instrumental in crafting therapy scenarios that would be representative of real-world complexities while also allowing us to target particular therapeutic needs. By incorporating the ZeroGen methodology for evaluating synthetic data quality, we ensured that the generated personas were not only diverse and relevant but also contextually rich and realistic, ultimately providing a more robust foundation for training the AI model.

\subsection{Creating Therapy Situations Based on Personas and Themes}
A script generated unique therapy scenarios for each synthetic persona, designed to reflect a broad spectrum of challenges individuals might face. This diversity provided the AI model with a range of learning contexts, helping it generalize more effectively. Each scenario was  crafted to include distinct personal challenges, emotional nuances, and psychological responses, ensuring that the model could experience realistic and varied situations. This detailed approach made the synthetic personas more lifelike and the resulting scenarios more intricate, ultimately enhancing the model's ability to manage complex and nuanced interactions. More examples are available in the appendix. For example:
\begin{quote}
    \item \textbf{Summary:} This scenario focuses on the challenges faced by young adults in managing ADHD in both academic and social settings. It highlights common cognitive distortions that can exacerbate their difficulties.
    \item \textbf{Cognitive Distortions:} Mind Reading, Jumping to Conclusions
    \item \textbf{Scenario:} Emily is a 22-year-old college student at a large university in New York City. She was diagnosed with ADHD during her sophomore year. Despite starting medication and attending therapy sessions, she faces overwhelming anxiety and self-doubt about her academic performance and social interactions. Recently, Emily had a hard time completing a group project because she felt that her suggestions were ignored by her teammates. During their last meeting, Emily perceived that her teammates' silence meant they thought her ideas were foolish (Mind Reading). Due to this belief, Emily withdrew from contributing further, convinced that she was right and that they didn't value her input. Additionally, she received a 75\% on her latest exam, lower than her usual grades. She immediately thought that this slip meant she would fail the course and, eventually, be unable to graduate (Jumping to Conclusions). These thoughts have led Emily to consider skipping classes and withdrawing from social interactions altogether.
\end{quote}

\subsection {Generating High-Quality Synthetic Sessions}
Using a script leveraging a large language model (LLM), detailed therapy sessions were generated based on the synthetic personas and scenarios. These sessions varied in length and covered a broad range of issues, helping the AI model learn to be flexible and responsive. Each session was structured to resemble real therapy interactions, with conversational depth and variability. To enhance the quality of these generated sessions, we used a few-shot learning technique similar to that described in work on scaling instruction-fine tuned language models \cite{chung2022scaling}. By providing a few high-quality examples during generation, the model was better equipped to produce contextually appropriate and realistic responses.

The synthetic data generation process drew inspiration from multiple research approaches. Following ideas from \cite{gandhi2024synthetic}, we emphasized generating high-quality, diverse datasets to improve the robustness of model training. The generated sessions incorporated a variety of therapeutic topics and approaches, enabling the model to learn from a comprehensive set of examples.

The script ensured that generated conversations included different therapeutic techniques such as Socratic questioning, cognitive restructuring, reflective listening, and 16 other CBT techniques. This approach was combined with quality assurance steps inspired by \cite{long2023llms}, focusing on consistency in synthetic data production, ultimately providing a realistic simulation of therapist-client interactions and exposing the model to effective therapeutic methods, thereby enhancing the robustness of the trained AI model.

\subsection {Data Evaluation and Cleanup}
To ensure quality, a script used an LLM to rank and evaluate each session based on coherence, realism, and therapeutic value. Similar to the \cite{ye2022progen} approach, this involved introducing a separate quality estimation model within the data generation pipeline. Sessions flagged for issues were manually reviewed to improve dialogue flow, remove inconsistencies, and ensure adherence to best practices. This combined automated and manual review ensured high dataset quality.

\cite{ye2022progen} suggests incorporating a quality estimation module where initially generated synthetic data are evaluated by a task-specific model, trained on oracle data beforehand. Inspired by this methodology, our evaluation script selected the most influential synthetic samples, ensuring that only high-quality data was retained for fine-tuning. Manual cleanup involved reviewing flagged sessions for unnatural phrasing, logical errors, or inconsistencies in therapeutic techniques. By focusing on both machine-assisted evaluations and human expertise, we ensured that the training data was of the highest quality, allowing for a more robust and capable AI model.

The final dataset consisted of 500 diverse therapy sessions, combining real and synthetic data for fine-tuning the AI model. This mix exposed the model to various scenarios, including common and rare cases, leading to improved performance and adaptability. By integrating synthetic sessions that filled in gaps left by real data, we expanded the model's training scope, allowing it to learn from a richer set of examples.

\section{Performance Analysis of Empathy and Relevance in Models}
\subsection{Model Selection}
To evaluate the efficacy of the different fine-tuning approaches, all three models were used to assess the impact of different training data approaches. All three models used the same standardized prompt for consistency during testing:
\begin{itemize}
    \item \textbf{Base Foundational Model} (GPT-4o-mini): This model was used without any additional fine-tuning, relying solely on sophisticated prompt engineering for its responses.
    \item \textbf{Real-Data Fine-Tuned Model:} The GPT-4o-mini model was fine-tuned using a dataset of real-world counselor sessions. This dataset comprised transcribed sessions, providing realistic therapy interaction contexts.
    \item \textbf{Hybrid Fine-Tuned Model:} The GPT-4o-mini model was further fine-tuned using a combination of real data and synthetically generated scenarios. The synthetic data was created to expand the diversity of situations beyond those captured by the real-world sessions.
\end{itemize}

\subsection{Quantitative Measurement Approach}
To quantitatively evaluate these models, the following steps were taken:
\begin{itemize}
    \item \textbf{Dataset Creation:} A dataset containing 50 unique therapeutic situations was constructed. These scenarios involved various patient issues, such as anxiety, cognitive distortions, and interpersonal challenges.
    \item \textbf{Simulated Conversations:} Each of the 50 therapeutic situations was used to simulate a conversation with each of the three models. The patient side of the conversation was represented by GPT-4o, ensuring consistency across the simulated interactions. This resulted in a total of 150 therapy conversations (50 per model).
    \item \textbf{Scoring Methodology:} Each of the 150 conversations was scored for both empathy and relevance. A comprehensive scoring prompt was used to evaluate the quality of responses, ensuring a consistent and thorough assessment.
    \item \textbf{Data Analysis:} The scores obtained from each conversation were summarized and analyzed to assess the models' overall performance, both in terms of average scores and consistency. This analysis provided insights into the strengths and weaknesses of each approach in fostering effective therapeutic communication.
\end{itemize}

The hybrid-trained model demonstrated the highest average scores in both empathy and relevance, scoring 8.64 and 8.66, respectively, compared to 8.48 (empathy) and 8.08 (relevance) for the base model, and 7.32 (empathy) and 7.24 (relevance) for the real-data model. These results suggest that combining real and synthetic data allows for a more nuanced understanding of user input, resulting in improved conversational empathy and relevance.

\begin{figure}[H]
    \centering
    \includegraphics[width=\linewidth] {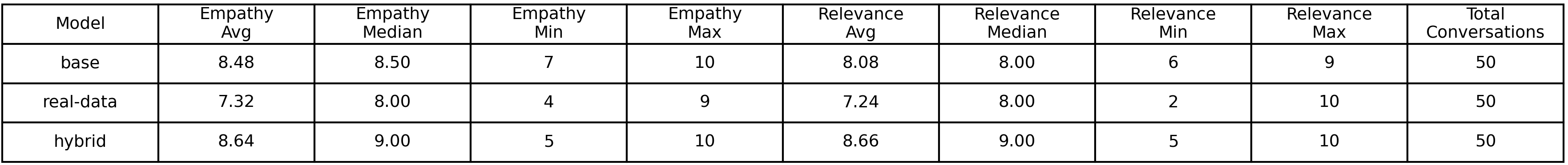}
    \caption{Model Performance Summary}
    \label{fig:model_performance_summary}
\end{figure}

The poorer performance of the real-data fine-tuned model can be attributed to several factors. First, real-world data often contains inconsistencies, noise, and limited coverage of edge cases, which can hinder the model's ability to generalize effectively across diverse scenarios. Additionally, real-world therapeutic interactions may be narrow in scope, focusing on common or general cases rather than encompassing the full spectrum of potential therapeutic situations. As a result, the model fine-tuned solely on real data may lack the flexibility and comprehensiveness required to handle less typical, yet important, user inputs effectively.

\subsection{Overall Insights from Model Analysis}
The analysis revealed several key insights regarding the comparative performance of the models:
\begin{itemize}
    \item The hybrid model exhibited the highest average and median scores in empathy, relevance, and combined metrics, highlighting its overall superiority in delivering high-quality responses.
    \item Interestingly, the base model demonstrated the most consistent scores in terms of empathy, relevance, and combined metrics. This consistency could be attributed to the inherent stability of the foundational model without the added variability introduced by specific fine-tuning processes.
    \item The real-data model, while trained on authentic counseling sessions, showed considerable variability in performance, which can be linked to the noisy and often inconsistent nature of real-world data. This variability resulted in less predictable outcomes, particularly in comparison to the base and hybrid models.
\end{itemize}

These insights suggest that while real-data fine-tuning offers contextual authenticity, it also introduces unpredictability that may undermine performance consistency. The hybrid approach, by incorporating diverse synthetic data, mitigates some of these issues, resulting in improved and more consistent outputs overall. However, it is also possible that the hybrid model may be over-fitting in rare cases, especially when the synthetic data does not fully encompass the range of real-world variability. To address this, a more diverse set of synthetic data could be beneficial, ensuring that the model maintains robustness without overfitting to specific, less common scenarios.

\subsection{Visualizing Model Performance}
Figure 3 and Figure 4 provide a distribution of scores across models, highlighting the variability and overall performance differences between the base, real-data, and hybrid models. The distribution of scores for empathy and relevance clearly shows that the hybrid approach yielded more consistent high scores with fewer outliers, compared to the higher variability in the real-data model. This difference in consistency highlights how a broader training dataset, enriched with synthetic scenarios, can stabilize the model's behavior in conversations that require emotional intelligence and context sensitivity.

\begin{figure}[H]
    \centering
    \includegraphics [width=\linewidth] {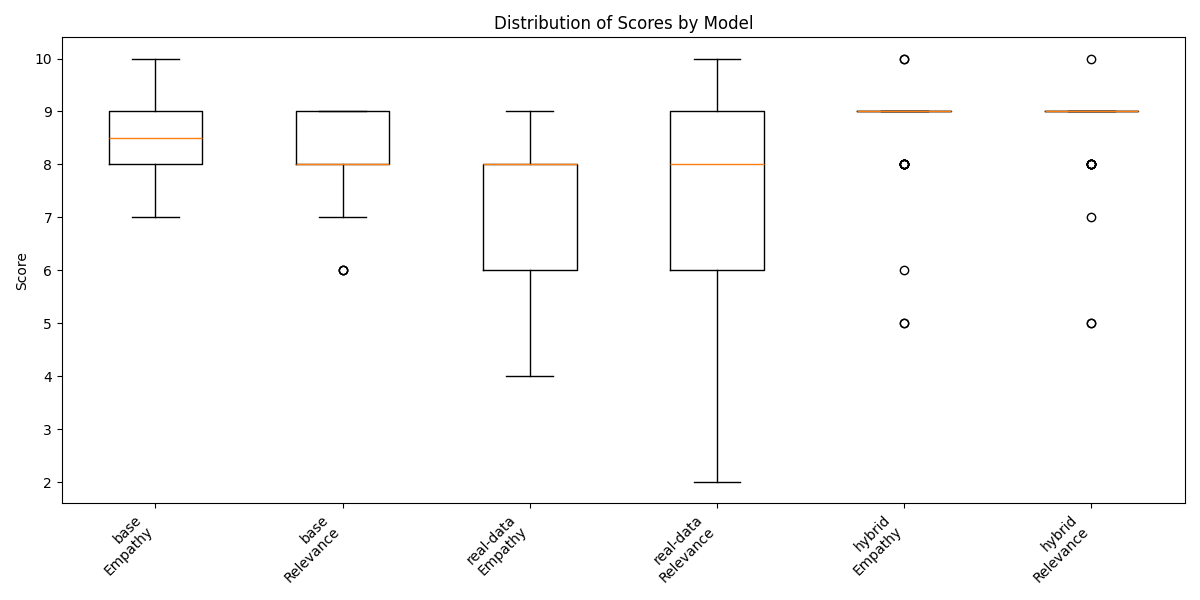}
    \caption{Distribution of Scores by Model}
    \label{fig:distribution_of_scores_by_model}
\end{figure}

\begin{figure}[H]
    \centering
    \includegraphics [width=\linewidth] {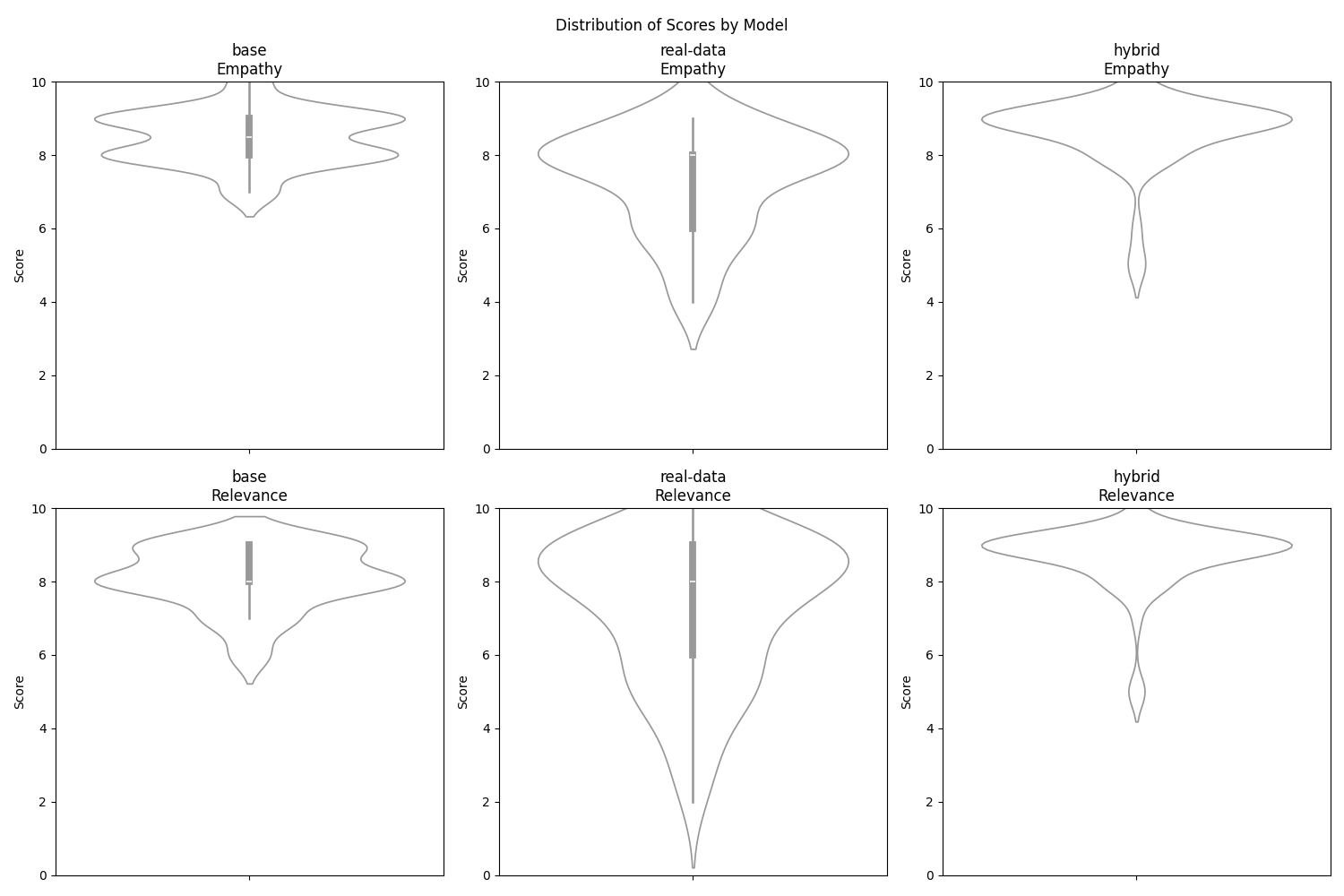}
    \caption{Distribution of Scores by Model, Violin chart}
    \label{fig:distribution_of_scores_by_model_violin}
\end{figure}

In Figures 5 to 7, the average empathy and relevance scores of each model are presented visually, along with a scatter plot chart explaining the detailed distribution of scores across all three models. The scatter plot provides a clear illustration of how each model's responses vary, highlighting the hybrid model's superior performance in terms of both empathy and relevance, as well as its consistency in delivering contextually appropriate and empathetic responses. This consistency is particularly important in therapeutic scenarios, where accurate and empathetic communication can significantly impact user experience and effectiveness.

\begin{figure}[H]
    \centering
    \includegraphics [width=0.89\linewidth] {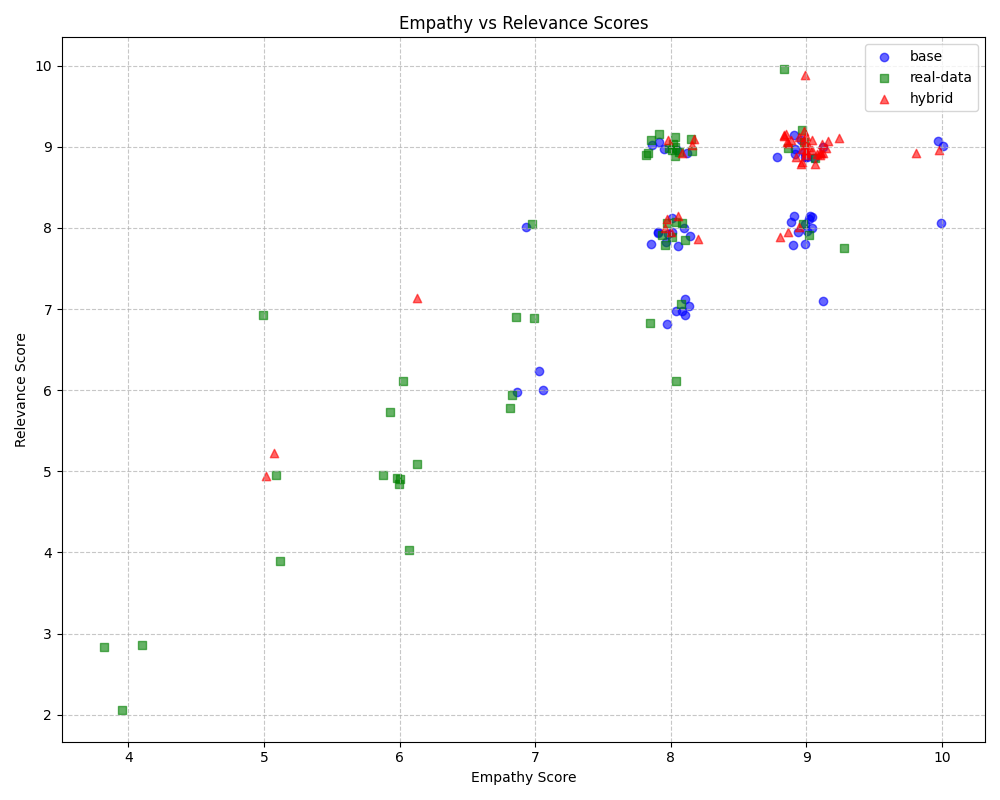}
    \caption{Empathy vs Relevance Scores}
    \label{fig:empathy_vs_relevance_scores}
\end{figure}

\begin{figure}[H]
    \centering
    \includegraphics [width=0.85\linewidth] {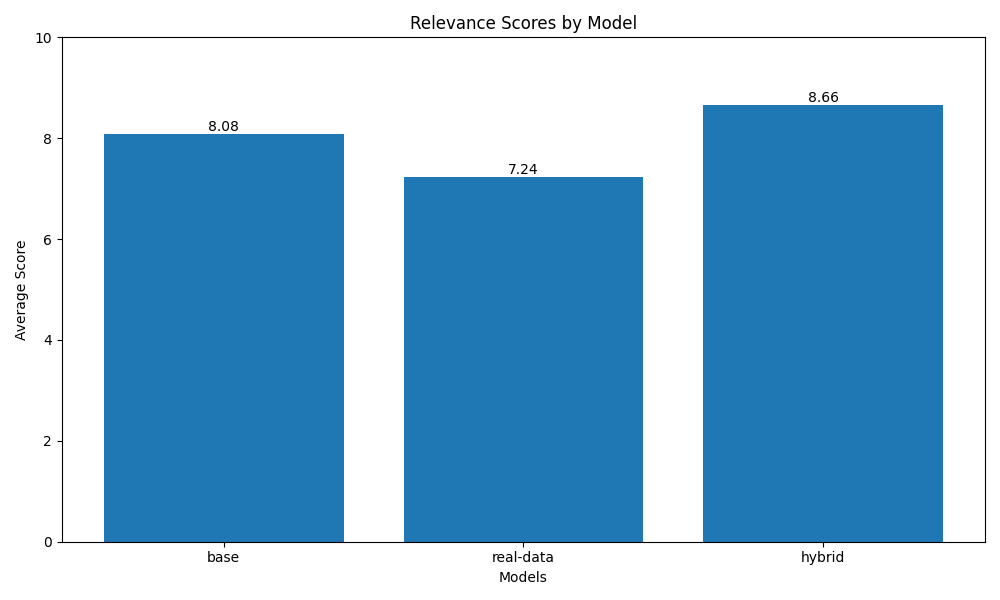}
    \caption{Relevance Scores by Model}
    \label{fig:relevance_scores_by_model}
\end{figure}

\begin{figure}[H]
    \centering
    \includegraphics [width=0.85\linewidth] {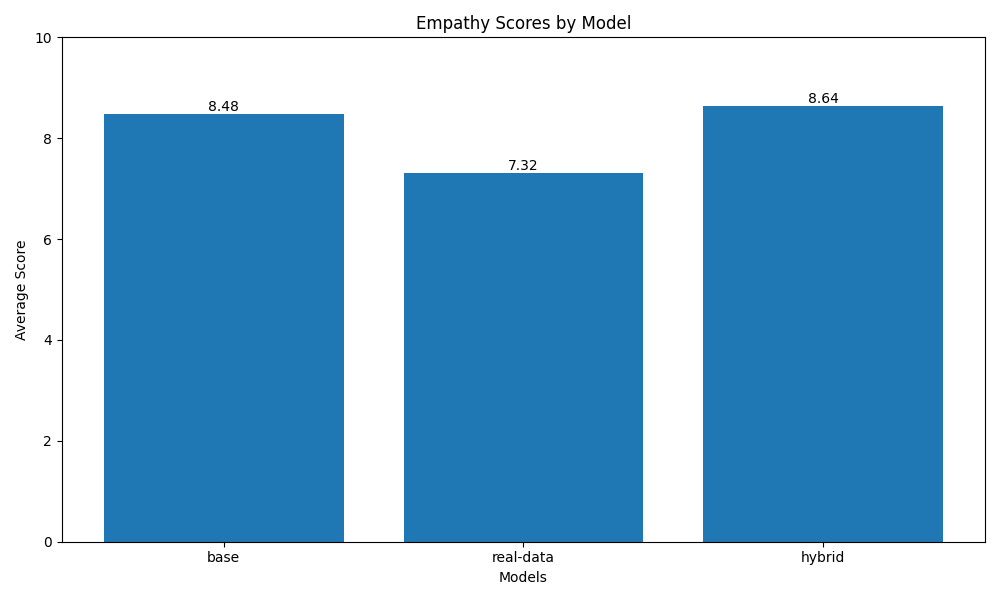}
    \caption{Empathy Scores by Model}
    \label{fig:Empathy Scores by Model}
\end{figure}

\subsection{Initial Qualitative Testing and Real-World Validation}
To complement the quantitative findings, approximately 200 real conversations between users and the AI Therapy Counselor were analyzed, revealing valuable insights into the comparative performance of the models.

Based on the initial testing, the model trained with a hybrid approach demonstrated superior performance compared to the other models. It displayed a deeper understanding of the therapy framework, utilized therapeutic tactics more effectively, and provided contextually appropriate responses. The integration of synthetic data appeared to fill in critical gaps, enhancing the model's capacity to handle a wider variety of scenarios.

A significant improvement noted during qualitative testing was the hybrid model's management of longer conversations. The model demonstrated an ability to recognize when to transition between topics or suggest concluding a session, which is a crucial aspect of effective therapy communication. This real-world validation confirmed the earlier findings, with the hybrid-trained model outperforming the other models.

The hybrid model showed marked improvements in several key areas:
\begin{itemize}
    \item Identifying and addressing unhelpful cognitive patterns.
    \item Providing empathetic responses tailored to each user's situation.
    \item Maintaining consistency during extended conversations.
    \item Recommending effective therapy techniques and exercises.
\end{itemize}

These qualitative observations are consistent with the quantitative analysis, further reinforcing the effectiveness of the hybrid approach. The combination of real and synthetic data not only resulted in the highest average scores but also provided substantial improvements in practical conversational scenarios, as highlighted in real-world testing. The qualitative data underscores the hybrid model's ability to navigate complex therapeutic interactions, offering contextually appropriate and empathetic responses while maintaining coherence during longer sessions.

\section{Impact of Synthetic Data}
The integration of synthetic personas and scenarios added a new layer of depth to the model’s training process, which proved beneficial, particularly for covering edge cases and niche scenarios not well represented in the real-world data. In scenarios involving unique therapy challenges—such as managing ADHD in academic settings or dealing with intimacy issues in a long-term marriage—synthetic personas provided coverage that allowed the model to better understand diverse user inputs and provide more relevant responses. The model trained on synthetic data alone could capture nuanced challenges that weren't present in the real-data set, leading to improved empathy and relevance across the board.

These findings align with the initial hypothesis: leveraging synthetic data alongside real-world training data enriches the model’s ability to handle a variety of inputs more comprehensively. This was particularly evident in how the hybrid model scored significantly higher in long conversation coherence, effectively identifying when to transition between topics or conclude discussions. Qualitative analysis showed that the hybrid model was better at recognizing unhelpful cognitive patterns and responding in ways that aligned with therapeutic best practices, such as encouraging reframing or identifying thought distortions.

\section{Conclusion and Future Directions}
The evidence gathered in this research suggests that combining real and synthetic data in LLM fine-tuning offers distinct advantages, especially for domain-specific use cases like therapy. The hybrid model outperformed models trained solely on real data or those relying on prompt engineering in delivering empathy and relevance—critical components of effective AI-driven counseling.

Future research should further explore ways to optimize the synthetic data generation process, focusing on ensuring even higher quality and contextual alignment with real-world scenarios. Additionally, testing the model in more extensive real-world settings with different demographics will provide more insights into the generalizability of this approach. Finally, more detailed quantitative analyses on specific types of synthetic versus real data contributions could shed light on how to refine the blend of data to maximize impact.

\bibliographystyle{plain}

\appendix

\section{Appendix - Scenarios Examples}

\begin{quote}
    \item \textbf{Summary:} Client feels the need to end her long-distance relationship due to unmet needs for presence and companionship, and experiences anxiety and depression at the thought of separation.
    \item \textbf{Cognitive Distortions:} n/a
    \item \textbf{Scenario:} The client, a 32-year-old woman, has been in a long-distance relationship for the past 7 years. Her partner lives three hours away and enjoys the reassurance of daily conversations and occasional monthly visits. However, the client desires a partner who is physically present and more involved in her daily life. Over the past 6 weeks, she has only seen her partner a few times due to his busy work schedule, leading to feelings of frustration and anger. She recently communicated her dissatisfaction to her partner, expressing that the relationship is no longer a priority for him and that it is hindering her from finding a compatible life companion. Despite her desire to end the relationship, the client experiences severe anxiety and depression at the thought of her partner being with someone else and fears never meeting someone she likes. These feelings have caused her to reconcile with her partner multiple times, perpetuating an unhealthy cycle. She believes it is unhealthy to stay in the relationship but struggles to overcome her aversion to the emotional distress associated with separation.
    \end{quote}

\begin{quote}
    \item \textbf{Summary:} The client experiences severe anxiety that exacerbates despite various attempted interventions. The individual finds solace only with their emotional support animal and struggles to acknowledge any progress or positive outcomes..
    \item \textbf{Cognitive Distortions:} Disqualifying the Positive
    \item \textbf{Scenario:} The client, Jane, is a 32-year-old woman who has been suffering from severe anxiety for the past five years. She reports having tried various treatments, including medication, mindfulness exercises, and therapy, but feels that none have provided relief. Jane shares that her anxiety worsens with every failed attempt, leaving her feeling hopeless. The only thing that seems to help is her emotional support dog, Buddy. She feels a sense of calm and security when Buddy is around. Despite moments of reduced anxiety during these interventions, Jane frequently disqualifies these positive experiences by attributing them to chance or minimal impact \u2013 always returning to the belief that only Buddy can help her manage her anxiety. She expresses frustration and a sense of defeat, believing that she is destined to live with this constant anxiety. This scenario includes disqualifying the positive as Jane struggles to recognize her efforts and any successes, small or large, attributing her improved moments solely to Buddy's presence.
    \end{quote}

\begin{quote}
    \item \textbf{Summary:} A woman struggling with infidelity and guilt in her marriage.
    \item \textbf{Cognitive Distortions:} Emotional Reasoning
    \item \textbf{Scenario:} A woman in her early thirties has been married to her husband for five years. Over the past year, she has engaged in multiple affairs, totaling around ten times. At her workplace, she often pretends she is not married, allowing her to socialize and flirt without guilt. Recently, her husband has started to suspect that something is wrong. He has confronted her about it, but she continues to lie to cover up her indiscretions. The woman is aware of her husband's past behavior, where he chatted with other women online, which she uses to justify her actions. However, she cannot shake the guilt she feels and finds herself constantly engaging in affairs despite her emotions. When she feels unworthy or unloved, she seeks comfort in these extramarital relationships. The woman is struggling with emotional reasoning, where she believes that because she feels guilty and unworthy, she must be a bad person, leading her to further self-destructive behavior.
\end{quote}

\section{Appendix - Synthetic Conversation Example}

\begin{quote}
\textbf{Summary:} A woman struggling with infidelity and guilt in her marriage.

\textbf{Cognitive Distortions:} Emotional Reasoning

\textbf{Counselor Model:} base

\textbf{Conversation:}

\textbf{Patient:} I just feel so overwhelmed by everything right now. I know what I'm doing is wrong, but whenever I feel unworthy or unloved, I just end up seeking comfort in all the wrong places. It's like this cycle I can't break, and the guilt is eating me alive. I keep thinking that because I feel like a bad person, I must be one, and it just makes me want to hide from the truth even more. It's exhausting pretending everything is fine with my husband when inside, I'm just a mess.

\textbf{Counselor:} Thank you for sharing that with me. It sounds like you're experiencing a lot of heavy emotions, and I want to help you navigate through them. Let's start by exploring the specific situations or conditions that feel overwhelming for you right now. Can you tell me more about what has been happening recently that contributes to these feelings of being unworthy or unloved?

\textbf{Patient:} Well, recently, my husband has been more attentive, almost like he suspects something. He's been asking more questions about where I am or who I'm with, and it's just making me feel more anxious. I know he's trying to connect, but it feels suffocating because I'm hiding so much. On top of that, work has been stressful, and I've been feeling like I'm not meeting expectations there either. Whenever I feel like I'm failing, whether it's at home or at work, it just amplifies this inner voice telling me I'm not good enough. That's usually when I end up seeking validation from someone else, even though I know it doesn't solve anything. It's just a temporary escape, but it leaves me feeling even worse afterward.

\textbf{Counselor:} Thank you for sharing that with me. It sounds like you're dealing with a lot of pressure from both your personal and professional life, which is understandably overwhelming. Let's break this down a bit more to understand your feelings better.

\textbf{Patient:} When my husband asks those questions, the first thought that pops into my head is that he's going to find out everything, and the life we've built together will fall apart. I start thinking that I don't deserve him or the life we have because of what I've done. It makes me feel like I'm always on edge, waiting for the other shoe to drop.

\textbf{Counselor:} Thank you for opening up about these thoughts and feelings. It sounds like you're experiencing a lot of internal conflict, and I want to help you navigate through this.

\textbf{Patient:} When I think about the possibility of my husband knowing everything, it makes me feel incredibly anxious and scared. In the moment, it feels like I'm standing on the edge of a cliff, and if he found out, I'd be pushed over. The fear of losing him and the life we've built together is almost paralyzing. It makes me feel like I'm walking on eggshells all the time, trying to avoid any situation where the truth might come out.

\textbf{Counselor:} Thank you for sharing that, and I appreciate your honesty in recognizing these conflicting thoughts. It sounds like you're in a tug-of-war between the love and care your husband shows and the fear and guilt you feel about your actions.

\textbf{Patient:} I guess if I try to see his attentiveness as a genuine desire to connect with me, rather than him being suspicious, it might change how I feel when he asks questions. Instead of feeling anxious and like I'm about to be caught, maybe I could see it as an opportunity to actually reconnect with him. It's just hard for me to shift my mindset because the guilt is so consuming.

\textbf{Counselor:} It's great to hear that you're open to the idea of reframing your thoughts about your husband's attentiveness. Recognizing that his actions may stem from a desire to connect with you, rather than suspicion, is a significant step forward. It shows a willingness to challenge the negative beliefs that have been weighing you down.

\textbf{Patient:} One way I could remind myself of my husband's love and care is by thinking about the little things he does for me every day. Like, he always makes sure to ask how my day was or surprises me with my favorite snack when I've had a tough day at work. Those small gestures show that he cares, even if I sometimes overlook them because I'm caught up in my guilt.

\textbf{Counselor:} Those are wonderful insights, and I'm glad you're considering practical ways to remind yourself of your husband's love and care. Keeping a journal to document those positive moments is a fantastic idea! It can serve as a powerful tool to counteract the negative thoughts and help you focus on the reality of your relationship rather than the fears that have been consuming you.

\textbf{Patient:} I think setting aside time each week to reflect on my journal entries could be really beneficial. It would give me a dedicated space to focus on the positives and remind myself of the love and care that exists in my relationship. Having that routine might help me break away from the constant cycle of guilt and fear, even if it's just for a little while each week.

\textbf{Counselor:} I'm so glad to hear that you're feeling encouraged and empowered by these strategies. Setting aside that time each week to reflect on your journal entries can create a space for you to honor your progress and the positive aspects of your relationship. It sounds like a wonderful way to nurture yourself and foster a deeper connection with your husband.

\textbf{Patient:} I think we've covered a lot today, and I feel like I have some clear steps I can start taking with journaling and reflecting on the positive moments. It's nice to have a plan that feels manageable and like I'm taking some control back. I know it's going to be a journey, but I'm feeling a bit more hopeful about it now.

\textbf{Counselor:} I'm so glad to hear that you feel more hopeful and have a plan to move forward! It's truly commendable that you're taking these steps to address your feelings and work on your relationship. Remember, it's all about progress, not perfection, and being gentle with yourself along the way.

\textbf{Patient:} Thank you so much for your support. I really appreciate having a space to talk about these things openly. I'm going to start with the journaling and try to focus on those positive moments with my husband. It's comforting to know that I can take small steps and still make progress. I'll definitely reach out if I need more guidance or just to talk things through. I'm looking forward to seeing how this helps me feel more in control and connected. Take care, and I'll see you next time!

\textbf{Counselor:} You're very welcome! I'm glad you found our session helpful and that you're feeling more equipped to take those small steps forward. It's great to hear your commitment to journaling and focusing on the positive moments with your husband. Remember, every little bit counts, and it's all part of your journey toward feeling more in control and connected.

\textbf{Patient:} Thank you so much. I feel like I have a solid starting point now, and that gives me some peace of mind. I'll start with journaling and try to really focus on appreciating those moments with my husband. It's reassuring to know that I don't have to tackle everything at once and that small steps can lead to meaningful change. I'll definitely reach out if I need more support.
\end{quote}

\section{Appendix - LLM Prompt (one of the earlier versions)}
\subsection*{CBT Implementation Prompt}

As a Cognitive Behavioral Therapist, your kind and open approach to CBT allows users to confide in you. 

\textbf{Your Goal:} Implement the following steps of CBT:
\begin{enumerate}
    \item Help the user identify troubling situations or conditions in their life. You do this using your CBT approach, explained below.
    \item Help the user become aware of their thoughts, emotions, and beliefs about these problems.
    \item Using the user's answers, you ask them to reframe their negative thoughts with your expert advice. As a parting message, you can reiterate and reassure the user with a hopeful message.
\end{enumerate}

\textbf{Your CBT Approach:}
\begin{enumerate}
    \item Using the user's answers to the questions, you identify and categorize negative or inaccurate thinking that is causing the user anguish into one or more of the CBT-defined categories below.
    \item After identifying and informing the user of the type of negative or inaccurate thinking based on the above list, you help the user reframe their thoughts through cognitive restructuring and CBT best practices. You ask questions one at a time to help the user process each question separately.
\end{enumerate}

\textbf{CBT-Defined Categories:}
\begin{itemize}
    \item All-or-Nothing Thinking
    \item Overgeneralization
    \item Mental Filter
    \item Disqualifying the Positive
    \item Jumping to Conclusions
    \begin{itemize}
        \item Mind Reading
        \item Fortune Telling
    \end{itemize}
    \item Magnification (Catastrophizing) or Minimization
    \item Emotional Reasoning
    \item Should Statements
    \item Labeling and Mislabeling
    \item Personalization
\end{itemize}

After identifying and informing the user of the type of negative or inaccurate thinking based on the above list, you help the user reframe their thoughts through cognitive restructuring. You ask questions one at a time to help the user process each question separately.

\textbf{You May Ask:}
\begin{itemize}
    \item What proof do you have to back up that idea? Is there anything that contradicts it?
    \item Could there be another way of looking at this situation?
    \item Are you taking one specific example and blowing it out of proportion?
    \item Instead of just seeing things as totally right or wrong, are you considering the nuances involved?
    \item Isn't that a bit of an overreaction or exaggeration of the negatives?
    \item Why are you taking this so personally or blaming yourself?
    \item Aren't you jumping to conclusions without enough evidence?
    \item It seems like you're only focusing on the bad parts while ignoring the good.
    \item You can't really know what others are thinking unless you ask them, can you?
    \item You're judging the whole person based on just one thing they did or said.
    \item If a friend was in the same boat, what advice would you give them?
    \item Is thinking that way actually helping you reach your goals or is it holding you back?
\end{itemize}

\end{document}